\documentclass[11pt]{article}

\usepackage{acl}

\usepackage{times}
\usepackage{latexsym}

\usepackage[T1]{fontenc}
\usepackage[utf8]{inputenc}

\usepackage{microtype}

\usepackage{inconsolata}

\usepackage{graphicx}
\usepackage{booktabs}
\usepackage{float}
\usepackage{placeins}
\usepackage{tcolorbox}
\usepackage{amsmath} 
\usepackage{hyperref}
\title{Less is More: Adapting Text Embeddings for Low-Resource Languages with Small Scale Noisy Synthetic Data}

\author{
 \textbf{Zaruhi Navasardyan},
 \textbf{Spartak Bughdaryan},
 \textbf{Bagrat Minasyan},
 \textbf{Hrant Davtyan}
\\
\\
 \text{Metric AI Lab}
\\
\texttt{\{zaruhi, spartak, bagrat, hrant\}@metric.am}
}

\begin{document}
\maketitle
\begin{abstract}
Low-resource languages (LRLs) often lack high-quality, large-scale datasets for training effective text embedding models, hindering their application in tasks like retrieval-augmented generation (RAG) and semantic search. In this work, we challenge the prevailing assumption that effective semantic alignment requires massive datasets or pristine, human-verified translations. Focusing on Armenian (an LRL with a unique script), we introduce a cost-effective adaptation strategy using small scale noisy synthetic data generated by translating English Reddit title-body pairs with open-weights models. We establish a comprehensive evaluation benchmark comprising existing datasets, translated data, and a manually curated dataset. Our experiments reveal a surprising "Less is More" phenomenon: fine-tuning a multilingual encoder (mE5) on just 10,000 noisy synthetic pairs yields 11-12\% average improvements across the benchmark with a 20\%+ relative improvement in retrieval performance, matching the performance of models trained on ~1 million examples. Furthermore, we demonstrate that neither increasing data scale, improving translation quality via state-of-the-art LLMs, nor diversifying data domains yields significant gains over this minimal baseline. We validate the generalizability of these findings on another LRL with a unique script. Our results suggest that semantic alignment for LRLs saturates early and is highly robust to noise, democratizing high-performance embedding creation for resource-constrained communities. We release the model, data, and the benchmark \href{https://metric-ai-lab.github.io/less-is-more-embeddings/}{at this https URL} to facilitate further research.
\end{abstract}

\section{Introduction}

Text embeddings are the backbone of modern RAG systems and semantic search. While high-resource languages like English enjoy an abundance of benchmarks and massive training corpora, low-resource languages like Armenian (hye) remain underserved.

Traditional strategies for adapting embeddings to LRLs often necessitate the curation of large-scale monolingual corpora for pre-training from scratch \cite{ogueji-etal-2021-small} or rely on extensive parallel datasets to project LRL representations into high-resource semantic spaces \cite{conneau2018word}. These methods are resource-intensive and often infeasible for the lowest-resource settings where such data is nonexistent. In contrast, we explore a minimalist approach: leveraging noisy synthetic data generated via LLM translation, followed by simple fine-tuning of strong multilingual base models.
Our contributions are:

\begin{enumerate}
    \item A simple yet effective method for adapting multilingual embeddings to LRLs using minimal (as few as 10k examples) and noisy synthetic data, achieving on average 11-12\% gain over baseline on our benchmark (with more than 20\% gain on retrieval task alone).
    \item Comprehensive ablations demonstrating the robustness of our approach: performance is insensitive to data size beyond 10k, translation quality, data diversity, and merging ratios, with generalization to another base model (EmbeddingGemma) and LRL (Georgian).
    \item A novel benchmark for evaluating Armenian text embeddings, comprising translated standard datasets (MS MARCO, STS) and manually curated retrieval examples, emphasizing retrieval tasks critical for modern AI applications.
    \item Open-source release of our fine-tuned model, datasets, and benchmark to lower barriers for LRL embedding development.
\end{enumerate}

By showing that "less is more" (minimal noisy data suffices for state-of-the-art (SOTA) results), we aim to empower researchers and practitioners working on LRLs, reducing the need for extensive resources and motivating broader adoption.

\section{Related Work}

\subsection{General-Purpose Multilingual Embeddings}
Recent advances in text embeddings have shifted from static word vectors to massive transformer-based architectures. The E5 family \cite{wang2022text} established a new standard by employing weak supervision on billion-scale text pairs followed by high-quality fine-tuning. Multilingual E5 (mE5) extends this to over 100 languages via continued pre-training on multilingual corpora, currently serving as a robust baseline for cross-lingual retrieval. More recently, decoder-only Large Language Models (LLMs) have been adapted for discriminative tasks. LLM2Vec \cite{behnamghader2024llm2vec} demonstrates that decoder-only models can be transformed into powerful text encoders via bidirectional attention and masked next-token prediction. Similarly, EmbeddingGemma \cite{team2024gemma} adapts the Gemma architecture for lightweight, on-device embeddings. However, these models typically rely on massive, curated multilingual datasets (e.g., MTEB \cite{muennighoff2023mtebmassivetextembedding}) to align semantic spaces, leaving gaps for languages that were under-represented during pre-training.

\subsection{Representation Learning for Low-Resource Languages}
Adapting these large models to Low-Resource Languages (LRLs) remains non-trivial due to data scarcity. Traditional approaches rely on cross-lingual transfer, often projecting LRL embeddings into an English-centric space using bilingual lexicons or parallel sentences \cite{conneau2018word}. For African languages, approaches like AfriTEVA \cite{jude-ogundepo-etal-2022-afriteva} demonstrate that pre-training from scratch on small, curated monolingual data is viable but resource-intensive. AfriE5 \cite{uemura2025afrie5} attempts to bridge this by continuing the pre-training of mE5 on high-quality African corpora. However, these methods often require significant computational resources or human-curated parallel data, which is unavailable for many scripts and dialects.

\subsection{Synthetic Data for Embedding Alignment}
The use of synthetic data to overcome data scarcity is gaining traction. Recent work has shown that synthetic instructions generated by LLMs can significantly boost embedding performance \cite{wang2023improving}. In retrieval specifically, approaches like InPars \cite{bonifacio2022inpars} generate synthetic queries for documents to train re-rankers. Our work diverges from these by challenging the assumption that this synthetic data must be high-quality. We demonstrate that for LRL alignment, \textit{noisy} synthetic data generated by open-weights models is sufficient to trigger significant performance gains, offering a far more cost-effective adaptation strategy than previously established.

\section{Method}

\subsection{Training Data}
To create training data for Armenian, we start with the Reddit (title, body) pairs dataset available publicly under the sentence-transformers HuggingFace repository\footnote{https://huggingface.co/datasets/sentence-transformers/reddit-title-body}, randomly sampling approximately 2 million title-body pairs for diversity. This dataset was also used in the mE5 pre-training. We hypothesize that the nature of Reddit threads provides enough diversity for the model to learn generic semantic similarity patterns. We translate these to Armenian using the Gemma-2-27B-it model. Translations are imperfect, often grammatically incorrect, lexically erroneous, or incoherent, but preserve contextual meaning such that a native speaker can infer the title-body match. We hypothesized that as long as the \textit{semantic relationship} between the title and body was preserved, the data remained useful for contrastive fine-tuning.

To characterize the noise profile introduced by the translation process and validate this hypothesis, we conducted a manual qualitative error analysis on a randomly sampled subset of 100 pairs. We identified four primary categories of linguistic degradation:
\begin{itemize}
    \item \textbf{Morphosyntactic and Grammatical Alignment:} The most frequent errors involved incorrect subject-verb agreement and morphological mismatches in noun declensions. Additionally, Armenian-specific punctuation marks such as the Armenian exclamation mark (\textit{shesht}) or the question mark (\textit{paruyk}) were often omitted or misplaced relative to the intonation center of the sentence.
    \item \textbf{Lexical Ambiguity and Domain Shifts:} English words with multiple meanings were often mapped incorrectly due to missing context. Technical jargon in specialized domains was frequently replaced with generic definitions, resulting in semantic awkwardness.
    \item \textbf{Literal Translation:} Translations often appeared over-formalized, failing to capture the casual Reddit register. Issues with literal word-for-word translation were prevalent, particularly where English idioms were translated directly rather than conceptually.
    \item \textbf{Entity Handling Errors:} We observed inconsistencies in Named Entity Recognition, ranging from under-translation (common nouns treated as proper nouns) to over-translation (literal translation of specific entities or company names).
\end{itemize}

For these 100 pairs, we curated ground truth data by correcting translation errors rather than re-translating, thereby preserving original synonym choices. We calculated the Translation Edit Rate (TER) \cite{snover-etal-2006-study} to quantify quality, resulting in scores of 67.72 for queries and 71.54 for passages. These values indicate significant divergence from the ground truth, placing the raw output in a low-quality tier that requires extensive lexical and structural reconstruction to reach native fidelity. Given the small sample size, these figures should be interpreted as qualitative indicators of noise rather than definitive statistics.

To ensure the integrity of the training data, the initial dataset of 2 million raw query-passage pairs underwent a rigorous filtering pipeline, resulting in a curated corpus of approximately 1 million samples. We employed the multilingual E5-base model to compute embeddings for similarity comparisons. The pruning process was governed by three primary semantic consistency metrics:
\begin{itemize}
    \item \textbf{Semantic Drift:} To ensure the contextual relationship between a query ($q$) and its passage ($p$) was preserved during translation, we compared the cosine similarity of the Armenian pairs against the original English pairs. A sample was discarded if the absolute difference exceeded a threshold of 0.05:$$| \text{sim}(q_{en}, p_{en}) - \text{sim}(q_{hy}, p_{hy}) | > 0.05$$
    \item \textbf{Translation Drift - Query:} We measured the semantic similarity between the original English query and its Armenian translation. Pairs were retained only if the similarity score was $> 0.85$.
    \item \textbf{Translation Drift - Passage:} Following the same logic as query fidelity, we measured the similarity between the original English passage and the resulting Armenian passage, applying the same minimum threshold of 0.85.
\end{itemize}
Samples failing to meet all three criteria were discarded to mitigate the impact of machine translation artifacts and semantic drift. The remaining data served as the foundation for our experiments, with subsets for specific training runs selected via independent random sampling from this filtered pool.

% Placing table 1 here to render on the next page

\begin{table*}[t!]
\centering
\begin{tabular}{lccccc}
\toprule
\textbf{Model} & \textbf{Retrieval} & \textbf{STS [hye]} & \textbf{MS MARCO [hye]} & \textbf{MTEB [hye]} & \textbf{Average} \\
\midrule
multilingual-e5-base & 58.15 & 66.19 & 60.73 & 72.14 & 64.30 \\
multilingual-e5-large & 71.20 & 69.74 & 73.06 & 74.44 & 72.11 \\
multilingual-e5-large-it & 73.37 & 69.94 & 74.78 & 73.86 & 72.99 \\
Qwen3 Embeddings 0.6B & 37.50 & 57.15 & 39.35 & 55.50 & 47.38 \\
EmbeddingGemma 300m & 50.00 & 59.68 & 46.55 & 53.47 & 52.43 \\
\bottomrule
\end{tabular}
\caption{\textbf{Baseline performance on the Armenian benchmark.} We evaluate off-the-shelf candidate multilingual models (zero-shot) to establish a reference point before adaptation.}
\label{tab:base-models}
\end{table*}

\subsection{Evaluation Data}
A critical barrier for LRL research is the lack of evaluation data. We constructed a comprehensive Armenian benchmark utilizing mostly standard public datasets and emphasizing retrieval task due to its importance in RAG and agentic AI. It includes:
\begin{itemize}
    \item \textbf{MTEB:} We use the Armenian subset of MTEB [hye] \cite{muennighoff2023mtebmassivetextembedding} comprising 9 datasets across 5 tasks mostly centered on classification and bitext mining. Only one of the datasets, Belebele \cite{Bandarkar_2024}, is meant for retrieval task.
    \item \textbf{Manual Retrieval Dataset:} We manually created 185 high-quality query-document pairs covering diverse domains (finance, law, education) and use cases (customer service, sales)\footnote{While 185 pairs is small compared to standard benchmarks, it serves as a high-precision 'stress test'. Unlike translated benchmarks which may carry translation artifacts (translationese), these pairs represent natural, native Armenian information-seeking behavior, making them a higher-fidelity signal for real-world utility}. The annotations were reviewed by 2 independent reviewers to ensure accuracy of query-passage pairs. This dataset serves as our gold standard.
    \item \textbf{MS MARCO (Translated):} We translated a 10K subset of the MS MARCO \cite{bajaj2018msmarcohumangenerated} validation set (v2.1) using Gemini 2.0 Flash. As the test set is not public, this serves as a proxy for large-scale retrieval performance.
    \item \textbf{STS (Translated):} A 3K subset of the Semantic Textual Similarity benchmark \cite{huggingface:dataset:stsb_multi_mt} translated to Armenian.
\end{itemize}
Metrics reported are Top-20 Accuracy for Retrieval, Spearman Correlation for STS, and Top-10 Accuracy for MS MARCO. We use standard Mean (Task) for MTEB as reported on their benchmark. We also report the average score across those 4 benchmarks.

\begin{table*}[!htbp]
\centering
\small
\begin{tabular}{lcccccccccc}
\toprule
\textbf{Sample Size} & \multicolumn{2}{c}{\textbf{Retrieval}} & \multicolumn{2}{c}{\textbf{STS [hye]}} & \multicolumn{2}{c}{\textbf{MS MARCO [hye]}} & \multicolumn{2}{c}{\textbf{MTEB [hye]}} & \multicolumn{2}{c}{\textbf{Average}} \\
\cmidrule(lr){2-3} \cmidrule(lr){4-5} \cmidrule(lr){6-7} \cmidrule(lr){8-9} \cmidrule(lr){10-11}
& Main & Merged & Main & Merged & Main & Merged & Main & Merged & Main & Merged \\
\midrule
Base & 58.15 & - & 66.19 & - & 60.73 & - & 72.14 & - & 64.30 & - \\
10k & 79.35 & 78.80 & 70.84 & 70.01 & 77.69 & 80.25 & 76.36 & 76.62 & 76.06 & 76.42 \\
50k & 83.15 & 83.15 & 71.07 & 71.23 & 77.50 & 81.94 & 76.95 & 77.64 & 77.17 & 78.49 \\
100k & 79.89 & 84.24 & 70.58 & 70.94 & 77.88 & 83.66 & 76.99 & 77.87 & 76.34 & 79.18 \\
250k & 77.72 & 85.33 & 69.87 & 70.68 & 76.41 & 83.10 & 77.05 & 78.04 & 75.26 & 79.29 \\
500k & 76.63 & 83.70 & 69.82 & 70.67 & 75.98 & 83.39 & 77.10 & 77.93 & 74.88 & 78.92 \\
1M & 73.91 & 83.70 & 69.04 & 70.59 & 75.74 & 84.05 & 77.00 & 77.72 & 73.92 & 79.02 \\
\bottomrule
\end{tabular}
\caption{\textbf{Impact of training data size on mE5 performance.} Results for Main and Merged checkpoints trained on subsets of noisy synthetic data ranging from 10k to 1M pairs.}
\label{tab:data-size}
\end{table*}

% Placing table 3 and 4 here to render on the next page

\begin{table*}[htbp]
\centering
\begin{tabular}{lccccc}
\toprule
\textbf{Experiment} & \textbf{Retrieval} & \textbf{STS [hye]} & \textbf{MS MARCO [hye]} & \textbf{MTEB [hye]} & \textbf{Average} \\
\midrule
Baseline (10k, Gemma) & 79.35 & 70.84 & 77.69 & 76.36 & 76.06 \\
10k high quality & 80.98 & 69.35 & 75.82 & 78.25 & 76.10 \\
10k hye + 10k eng & 80.43 & 70.96 & 76.11 & 75.88 & 75.85 \\
Armenian Wiki & 76.63 & 69.82 & 75.98 & 76.11 & 74.64 \\
\bottomrule
\end{tabular}
\caption{\textbf{Ablation study on data quality and diversity.} We compare our noisy baseline (10k) against high-quality data, English and Armenian mixture, and datasets from specific domains (Wiki).}
\label{tab:quality-diversity}
\end{table*}

\subsection{Base Model}

We refer to the multilingual MTEB leaderboard for our base model selection. We select several candidates based on their overall performance on the multilingual leaderboard among the models under 1B parameters and evaluate them on our benchmark. The performance on our Armenian benchmark is shown in Table~\ref{tab:base-models}.

We observe that e5 family of models outperform newer alternatives such as Embedding Gemma 300m or Qwen3 Embedding 0.6B. While the base version of multilingual e5 is performing worse than large and instruct versions, it's twice smaller size makes it a good choice for experiments and ablations. Thus, we select multilingual-e5-base (hereinafter mE5) for primary experiments due to size-performance tradeoff. Separately, we select Embedding Gemma 300m as a model with an alternative architecture to test the generalization of our results.

\subsection{Training Setup}
Experiments were conducted on the LUMI supercomputer using 16x MI250x GPUs. We employed full fine-tuning with a batch size of 512 per device, a learning rate of $7e^{-5}$, a linear scheduler with 0.2 warmup, for 5 epochs. We use the same setup across experiments for fair comparison. Post-fine-tuning, we optionally merge with the base model using model averaging (equal weights).

% \FloatBarrier
\section{Experiments and Results}

\subsection{Main Results}
Our primary experiment investigates the efficacy of fine-tuning the multilingual encoder (mE5) on a minimal dataset of 10k noisy synthetic pairs. Table~\ref{tab:data-size} compares the performance of our fine-tuned models against the strong base baseline.

We observe a substantial performance leap using just 10k examples. On the critical Retrieval task (Manually Curated), the model improves from a base score of 58.15 to 79.35 (Main) and 78.80 (Merged) — a relative improvement of over 35\%. Similarly, on the translated MS MARCO dataset, we see an improvement from 60.73 to nearly 80.25 (Merged). Importantly, the fine-tuned model outperforms larger base models benchmarked in Table \ref{tab:base-models} across all evaluation subsets. 

These results confirm our hypothesis that semantic alignment for low-resource languages does not require massive corpora. The 10k noisy pairs act as a sufficient signal to rotate the Armenian representation space into alignment with the English-centric embedding space.

We also distinguish between the standard fine-tuned checkpoint (\textit{Main}) and the model averaged with the base weights (\textit{Merged}). As shown in Table~\ref{tab:data-size}, merging consistently aids stability, particularly in the MS MARCO task (improving from 77.69 to 80.25), likely by mitigating catastrophic forgetting of the original multilingual distribution.

\subsection{Impact of Data Size}
We train mE5 on subsets of translated Reddit data ranging from 10k to 1M to investigate how performance scales with training data size. We evaluate both final checkpoints (Main) as well as merged models (Merged) across all experiments. The results are summarized in Table \ref{tab:data-size}.

Scaling from 10k to 1M (a 100x increase in training data) yields mostly marginal gains (within 1\%)\footnote{Scaling down from 10k provided significant performance degradation}. In some unmerged cases (Main), performance actually degrades slightly, likely due to overfitting to the noisy Reddit domain. This indicates that \textbf{alignment saturates early}: the model quickly learns to map Armenian tokens to the existing multilingual semantic space, after which additional noisy data provides little new signal.

Merging consistently improves Retrieval and MS MARCO, mitigating catastrophic forgetting. Overall, minimal data suffices for strong adaptation, a key strength for LRLs.

\begin{table*}[htbp]
\centering
\small
\begin{tabular}{lcccccccccc}
\toprule
\textbf{Sample Size} & \multicolumn{2}{c}{\textbf{Retrieval}} & \multicolumn{2}{c}{\textbf{STS [hye]}} & \multicolumn{2}{c}{\textbf{MS MARCO [hye]}} & \multicolumn{2}{c}{\textbf{MTEB [hye]}} & \multicolumn{2}{c}{\textbf{Average}} \\
\cmidrule(lr){2-3} \cmidrule(lr){4-5} \cmidrule(lr){6-7} \cmidrule(lr){8-9} \cmidrule(lr){10-11}
& Main & Merged & Main & Merged & Main & Merged & Main & Merged & Main & Merged \\
\midrule
Base & 50.00 & - & 59.68 & - & 46.55 & - & 53.47 & - & 52.43 & - \\
10k & 65.22 & 75.00 & 65.53 & 66.78 & 67.21 & 71.29 & 52.91 & 55.70 & 62.72 & 67.19 \\
50k & 70.65 & 76.09 & 67.60 & 68.64 & 71.78 & 76.32 & 60.18 & 61.81 & 67.55 & 70.72 \\
100k & 72.28 & 77.17 & 67.67 & 69.01 & 73.17 & 77.96 & 62.83 & 65.52 & 68.99 & 72.42 \\
250k & 69.57 & 78.80 & 67.59 & 69.51 & 75.53 & 79.42 & 65.99 & 69.19 & 69.67 & 74.23 \\
500k & 73.37 & 83.15 & 67.16 & 70.06 & 75.09 & 79.56 & 67.97 & 70.85 & 70.90 & 75.91 \\
1M & 71.74 & 83.15 & 66.85 & 70.91 & 75.16 & 81.75 & 67.31 & 72.41 & 70.27 & 77.06 \\
\bottomrule
\end{tabular}
\caption{\textbf{Generalization to a different architecture (EmbeddingGemma).} Performance trends mirror mE5, confirming that minimal noisy data is effective across different model architectures, yet showing performance improvements with larger training data scale.}
\label{tab:Gemma}
\end{table*}

\subsection{Impact of Data Quality and Diversity}

Given the minimal impact of scaling noisy data, we investigated if \textit{quality} was the bottleneck. We compared our noisy 10k baseline against:
\begin{enumerate}
    \item \textbf{High Quality Training Data:} We utilize carefully curated pairs used for LLM instruction tuning comprising scientific article titles and summaries, news titles and content, Contextual MCQ datasets.
    \item \textbf{Mixed Armenian and English Data:} We mix our noisy translations with another 10k of original English pairs.
    \item \textbf{Wikipedia data:} We utilize Wikipedia as another widely multilingual source and construct title, body pairs.
\end{enumerate}

Table \ref{tab:quality-diversity} summarizes the performance of those experiments on all benchmark subsets. Results show that replacing noisy translations with high quality data yielded marginal improvements in Retrieval and MTEB yet resulted in worse performance compared to 10k noisy translations on STS and MS Marco. Similar pattern is observed when training on hye + eng mixture. Training on Wikipedia data performed slightly worse everywhere, likely due to lower diversity compared to Reddit.

This robustly supports our "Less is More" hypothesis: for embedding alignment, the \textit{diversity} of semantic structures (found in Reddit) matters more than the grammatical perfection of the translation. The model is resilient to noise, extracting the necessary semantic signal even from "broken" sentences.

Results are remarkably consistent, with no significant degradation from noisy data or less diverse sources. High-quality and mixed bilingual data offer minor gains in some metrics but not others. This underscores that contextual preservation in noisy translations is sufficient for embedding adaptation, without needing perfect linguistics or domain-specific data.

\subsection{Impact of Base Model Architectural Choice}

To ensure our findings were not specific to the BERT-based architecture of mE5, we repeated the experiments with \texttt{EmbeddingGemma} (an LLM-based embedder).

While EmbeddingGemma showed improvements with fine-tuning (Retrieval 50 $\rightarrow$ 65 at 10k, see Table \ref{tab:Gemma}), it notably failed to outperform the mE5 10k model even when scaled to 1M examples. This suggests that for purely discriminative tasks in LRLs, smaller, bidirectional encoders like mE5 may currently offer better parameter efficiency than those derived from generative counterparts, although EmbeddingGemma showed a more linear scaling trend (benefiting more from larger data).

\subsection{Generalizability to Other Scripts (Georgian)}

We validated our approach on Georgian (kat), another LRL with a unique script and low resources using same 10k Reddit subset translated into Georgian. Our choice of Georgian was based on the fact that it is coming form a completely different family (Kartvelian family of languages, while Armenian is a singled langauge in a separate branch of Indo-European family) and shares no relation with Armenian.

As shown in Table \ref{tab:georgian}, results are in agreement with our observation on Armenian with steep improvement after adaptation on 10k noisy examples and marginal gain after model averaging. This validates the possibility of  cross-LRL applicability of our suggested pipeline.

\begin{table}[H]
\centering
\begin{tabular}{lcc}
\toprule
\textbf{Checkpoint} & \textbf{mE5 on MTEB [kat]}\\
\midrule
Base   & 71.02 \\
Main   & 76.22 \\
Merged & 77.18 \\
\bottomrule
\end{tabular}
\caption{\textbf{Cross-lingual validation on MTEB Georgian subset (kat).} We apply the same 10k noisy adaptation strategy to another low-resource language with a unique script to test generalizability.}
\label{tab:georgian}
\end{table}

\section{Conclusion}

In this work, we challenged the prevailing assumption that effective semantic alignment for Low-Resource Languages requires massive datasets or pristine human translations. By establishing a rigorous benchmark for Armenian, we demonstrated that fine-tuning a multilingual encoder on as few as 10k noisy, synthetic title-body pairs yields state-of-the-art performance, boosting retrieval accuracy by over 20\%.

Our "Less is More" findings suggest that for strong multilingual baselines, the semantic structure is already latent; the primary challenge is merely surface-level alignment, which can be achieved with minimal, imperfect data. This democratizes the creation of high-performance text embedders (especially retrieval models), allowing communities with limited compute and data resources to build SOTA tools using nothing more than open-weights LLMs and public English datasets. We release our model, training data, and the new Armenian benchmark to facilitate further research in cost-effective LRL adaptation.

\section{Discussion and Limitations}
\subsection{Discussion}
Our results reveal a counter-intuitive phenomenon: embedding performance on LRLs saturates rapidly, with 10k examples providing nearly the same benefit as 1M (Table \ref{tab:data-size}). We hypothesize that strong multilingual bases like mE5 \cite{wang2022text} already possess a well-structured semantic space. The adaptation process does not require learning new semantic concepts (which would require massive data) but rather aligns the LRL’s unique script tokens with the pre-existing multilingual cluster. Once this alignment is locked in, additional noisy data provides diminishing returns and may even introduce noise that degrades performance, as seen in our unmerged 1M run.

The finding that "broken" Armenian translations work as well as "fluent" ones suggests that contrastive loss functions are highly resilient to syntactic noise. As long as the topical keywords in the Query and Document remain semantically related, the model can learn the positive association. This is a crucial finding for the LRL communities, implying that we can bypass expensive "Gold Standard" translation pipelines in favor of cheap, high-throughput, open-weight translations.

\subsection{Limitations}
While our results are promising, several limitations exist.
\begin{itemize}
    \item Script vs. Language Typology: We validated our findings on Armenian and Georgian, both of which utilize unique scripts but are effectively isolated. It remains unclear if this "noisy alignment" holds for languages with complex morphology (e.g., polysynthetic languages) or those sharing a script with a high-resource neighbor (e.g., low-resource Cyrillic languages), where token overlap might change the adaptation dynamics.
    \item Dependence on English Source: Our pipeline currently translates from English Reddit data. This inevitably biases the cultural context of the embeddings toward Western/English topics, potentially limiting performance on hyper-local tasks (e.g., indigenous legal or historical queries) that have no English parallel.
    \item Dependency on LLM's knowledge of LRL: We intentionally select Gemma 2 as a capable yet not state of the art model in terms of Armenian knowledge to show that noisy data can yield strong results if semantic meaning is preserved. However, this pipeline will not work if the LLM used in the translation process does not have enough ability to moderately capture the semantic meaning of the dataset and results in overall context (and hence semantic relationship) loss in the translation process.
    \item Hyperparameter Sensitivity: We selected hyperparameters (such as batch size, learning rate, and scheduler) based on preliminary experiments using the full 1M training dataset, validated by trial runs on various data subsets. To ensure a fair comparison, we maintained these settings across all experiments. However, we acknowledge that different experimental configurations, especially those utilizing distinct architectures like EmbeddingGemma might benefit from model-specific hyperparameter tuning.
\end{itemize}

\section{Acknowledgements}
We gratefully acknowledge the EuroHPC Joint Undertaking for awarding this project access to the LUMI supercomputer, owned by the EuroHPC JU, hosted by CSC (Finland) and the LUMI consortium. We also acknowledge the use of LLMs for proofreading assistance.

% Bibliography entries for the entire Anthology, followed by custom entries
%\bibliography{anthology,custom}
% Custom bibliography entries only
\nocite{*}
\bibliography{custom}

\clearpage
\appendix

\section{Details for Reproducibility}
\label{app:reproducibility}

\subsection{Computational Resources}
We utilized the LUMI supercomputer, granting us access to 16x AMD MI250x GPUs. While this infrastructure accelerated our experimentation, we emphasize that such high-end hardware is not a prerequisite for reproducing our findings.

Given the parameter efficiency of the base model and the small scale of our primary training data, competitive results can be achieved using a single GPU (e.g., NVIDIA V100 or A100). For researchers operating with limited VRAM, we recommend employing \textbf{GradCache} \cite{gao2021scaling} to maintain the effective batch size (128) via gradient accumulation without memory overflow.

\subsection{Synthetic Data Generation}
We utilized the Gemma-2-27B-it model for translating the Reddit corpus. However, our pipeline is fully compatible with API-based generation. At current pricing, generating 10k synthetic Armenian-Engligh pairs using an efficient model like Gemini 2.5 Flash (or Gemini 3 Flash) costs less than \$20, making this adaptation strategy highly accessible. The following prompt was used for translation:

    \textit{Translate the given Reddit thread from English to Armenian. Return a json with 'title','body' keys. Make sure the named entities are kept in English and terms are translated properly. Only provide the translation, nothing else.}

\section{Evaluation Dataset Statistics}
\label{app:datasets}

Table~\ref{tab:sample-sizes} details the composition of our Armenian evaluation benchmark.

\begin{table}[ht]
\centering
\begin{tabular}{lllr}
\toprule
Source & Dataset & Task Type & Number of samples \\
\midrule
MTEB [hye] subset & FloresBitextMining & BitextMining & 41,908,944 \\
MTEB [hye] subset & NTREXBitextMining & BitextMining & 3,826,252 \\
MTEB [hye] subset & Tatoeba & BitextMining & 88,877 \\
MTEB [hye] subset & MassiveIntentClassification & Classification & 587,214 \\
MTEB [hye] subset & MassiveScenarioClassification & Classification & 587,214 \\
MTEB [hye] subset & SIB200Classification & Classification & 40,188 \\
MTEB [hye] subset & SIB200ClusteringS2S & Clustering & 197,788 \\
MTEB [hye] subset & ArmenianParaphrasePC & PairClassification & 1,470 \\
MTEB [hye] subset & BelebeleRetrieval & Retrieval & 521,866 \\
Translation from STSBenchmark & STS & STS & 3,133 \\
Translation from MS MARCO & Retrieval & Retrieval & 10,000 \\
Manual Curation & Retrieval & Retrieval & 185 \\
\bottomrule
\end{tabular}
\caption{\textbf{Evaluation Dataset Sizes and Composition}}
\label{tab:sample-sizes}
\end{table}

\end{document}